\documentclass[12pt,twocolumn]{article}

\usepackage{times}
\usepackage{latexsym}
\usepackage{amsmath}
\usepackage{multirow}
\usepackage{url} 

\usepackage{graphicx}

\usepackage{float}
\usepackage{lingmacros}
\usepackage{tree-dvips}
\usepackage{graphicx}
\usepackage{supertabular}
\usepackage{array}
\usepackage[english]{babel}

\usepackage{multicol}

\usepackage[font=small,labelfont=bf]{caption}

\begin{multicols}{2}
\title{Evolving Shepherding Behavior with Genetic Programming Algorithms}
\author{
Joshua Brul\'{e} \\
Department of Computer Science \\
University of Maryland \\
{\tt jtcbrule@gmail.com} \\
\and
Kevin Engel \\
Department of Computer Science \\
University of Maryland \\
{\tt kevin.t.engel@gmail.com} \\
\and
Nick Fung \\
Department of Computer Science \\
University of Maryland \\
{\tt nfung13@gmail.com} \\
\and
Isaac Julien \\
Department of Computer Science \\
University of Maryland \\
{\tt ijulien6@gmail.com} \\
 }
\date{}

\begin{document}
\maketitle
\end{multicols}
\begin{abstract}
We apply genetic programming techniques to the `shepherding'
problem, in which a group of one type of animal (sheep dogs)
attempts to control the movements of a second group of animals
(sheep) obeying flocking behavior.
Our genetic programming algorithm evolves an expression tree that governs the movements of each dog. The operands of the tree are hand-selected features of the simulation environment that may allow the dogs to herd the sheep effectively. The algorithm uses tournament-style selection, crossover reproduction, and a point mutation.
We find that the evolved solutions generalize well and outperform a (naive) human-designed algorithm.
\end{abstract}



\section{Introduction}

While flocking is a popular topic, algorithms for shepherding are less well-studied. Existing approaches to shepherding typically train a predictive model as in \cite{sumpter}, or employ predefined strategies which may be combined to achieve a
goal \cite{bennet}. Typically, the goal of shepherding is to herd the flock to some location which we will refer to as the `pen'.

In our approach, the shepherding system has no predefined strategies nor predictive modeling. Using standard genetic programming techniques, we evolve the expression tree for a pure (stateless) function which acts as the `force update' for each simulated sheep dog at each time step. The parameters of this function can include some combination of the dog's position, the position of the other dogs (in cooperative herding), the position of the nearest `free' (uncaptured) sheep, the center of mass of the flock, and a `steering point,' a position such that the line between the steering point and entrance to the pen crosses some sheep \cite{lien}.

Our simulated sheep all obey the same simple, fixed flocking rules designed such that the sheep attempt to cluster with each other and avoid the sheep dog(s). The fitness of the evolved `dog-AI' is the fraction of the sheep `captured' after some fixed number of simulation steps.

\section{Related Work}

(Lien et. al, 2005) studies shepherding behavior in an environment with multiple shepherds cooperating to control a flock \cite{lien}.
Shepherds, which exert a repulsive force on the flock, must find steering points to influence the direction of the flock as
desired. The steering points for the group of shepherds form either a line or an arc on a side of the flock, and each shepherd
chooses a steering point to approach based on one of several proposed heuristics.

(Sumpter et. al, 1998) presents a machine vision system that models the position and velocity of a flock of animals \cite{sumpter}.
A Point Distribution Mode is used to generate features based on input from a camera mounted on a "Robotic Sheepdog,"
and these features are then used to estimate a probability distribution of the movement of the flock over time, conditional on its
previous locations and velocities. This probability distribution is estimated using competitive learning in a neural network.
Finally, the robot can herd a flock of animals toward a goal by a maximum likelihood estimate of the robot's own path.

(Bennet and Trafankowski, 2012) provides an analysis of flocking and herding algorithms, and also introduces a
herding algorithm based on specific strategies inspired by real sheepdogs\cite{bennet}.
\cite{bennet} also considers using one of several flocking strategies for the animals being herded, and finds
that the success of different a herding algorithm is often dependent on the flocking behavior.

(Cowling and Gmeinwieser, 2010) uses a combined top-down and bottom-up approach to provide realistic sheep herding
in the context of a game. A finite state machine associated with each sheepdog represents possible herding strategies,
such as circling, and the state of the FSM is controlled at the top level by an AI "shepherd" \cite{cowling}.


\section{Simulation}

\subsection{Environment}
The simulation environment takes place within an enclosed, square area designated as the field. The field boundaries can be thought of as fences, which act as barriers to both sheep and dogs.  If an agent attempts to pass through a fence, the penetrating coordinate is reset to the fence coordinate and the corresponding velocity is set to zero.

\begin{figure}[H]
\includegraphics[width=\linewidth]{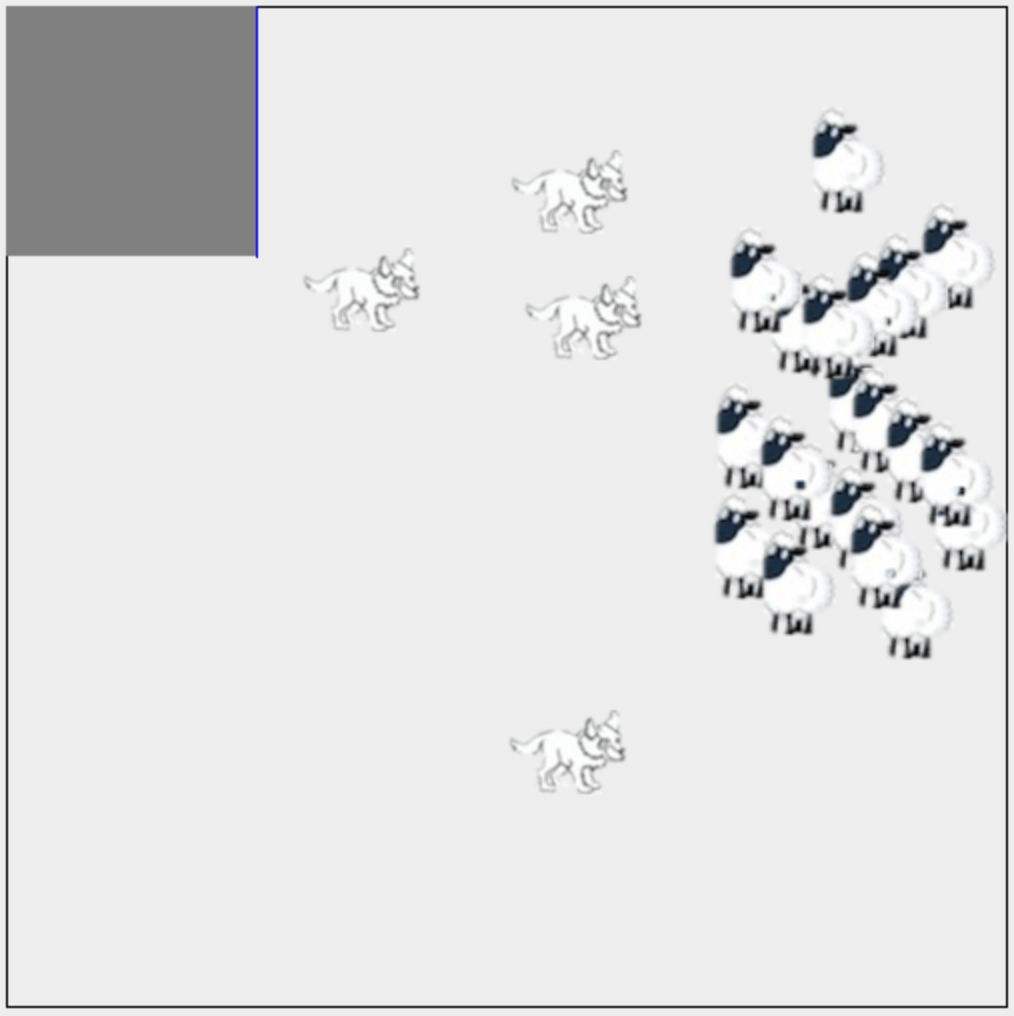}
\caption{Screenshot of Simulation Environment}
\label{simulator}
\end{figure}

The goal of the sheep dogs are to move the sheep into the pen, located in the top left corner of the field. Only the bottom side of the pen is open to the field, with the fence on the right side of the pen present to increase the difficulty of the task and help prevent sheep from randomly drifting in.

\subsection{Agents}

The simulation agents consist of sheep and dogs. The sheep all follow the same fixed rules, updating their position based on a set of force vectors:

\emph{Sheep clustering force:} Sheep which are within distance $d_s$ of one another are attracted/repelled with magnitude:
\begin{equation}
F_a - F_r\left(\frac{d_s^2}{|\vec{x}_{s_1}-\vec{x}_{s_2}|^2}-1\right)\,.
\end{equation}

\emph{Dog avoidance force:} Sheep which are within distance $d_d$ of a dog are repelled with magnitude:
\begin{equation}
F_d\left(\frac{d_d^2}{|\vec{x}_s-\vec{x}_d|^2}-1\right)\,.
\end{equation}

\emph{Fence avoidance force:} Sheep which are within distance $d_f$ of a fence are repelled with magnitude:
\begin{equation}
F_f\frac{d_f-d_\perp}{d_f}\,,
\end{equation}
where $d_\perp$ is the perpendicular distance between the sheep and the fence.

The above forces apply to free sheep moving in the open field area.  Upon entering the pen, a sheep is captured - its position is fixed to the corner of the pen and is no longer updated, effectively removing the sheep from the simulation.

The dogs, on the other hand, are controlled by an evolved genetic program (the `dog AI') which outputs a force for each dog at each time step in the simulation. The only constraint on the dog behavior is that they, like the sheep, are not allowed to penetrate fences, as described in the previous section.

\subsection{Initial conditions and simulation parameters}
To encourage the evolution of general solutions to the shepherding problem, random initial conditions were used for the agents.  Dogs spawn randomly inside the pen with no initial velocity, and sheep spawn randomly in the right half of the field with random initial velocities.  The right half of the field was chosen to minimize the possibility of sheep wandering into the pen without guidance from the dogs.

Each simulation was run with $\delta t = 1$ for $500$ time steps, returning the number of captured sheep at the end.  The field and pen were chosen to be squares of sizes $100 \times 100$ and $25 \times 25$ respectively.  For the majority of the results discussed in this paper, the following parameters were used:
\begin{align}
F_a &= .1\,,\,F_r = .05\,,\,F_d = 5\,,\,F_f = 1\,,\nonumber\\
d_s &= 20\,,\,d_d = 30\,,\,d_f = 5\,,\nonumber\\
v_s &= 1\,,\,v_d = 3\,.
\end{align}
The last two parameters are maximum allowed velocities - if a force increases a velocity past the maximum, it is rescaled to $v_s$ for sheep or $v_d$ for dogs.


\section{Genetic Program}

\subsection{Expression Trees}
A `dog AI' is a pair of pure (memoryless) functions which act on a set of parameters passed in by the simulation.  We experimented with several different parameter sets, detailed in Section \ref{Experimentation}.  The functions are stored as s-expressions (i.e. trees) whose leaves are either one of the parameters of the function, or some random number (drawn from a normal distribution).

Since updating the dog's position requires two values ($F_x$ and $F_y$), the root node of the tree is always `pair', with two subtrees representing the logic for the $x$ and $y$ forces, respectively. The internal nodes of a `dog AI' expression tree are one of the following operators: $-$ (unary), $-$ (binary), $+$, $*$, $div$ (where div is \emph{protected} division; attempting to divide by zero will return 1 instead).

In addition to these basic mathematical operations, we included qif(a, b, c, d), `quaternary if', which evaluates to $c$ if $a \leq b$ and $d$ otherwise. Note that all functions in the expression trees are functions of real numbers (implemented as JVM doubles), guaranteeing closure under the genetic operators .

\subsection{Genetic Operators and Initialization}
The design of the genetic programming system closely followed standard practice as described in \cite{IEC}. We used generational GP with a fixed population size $P$ initialized using the \emph{ramped half-and-half} method; half of the initial population was fully grown to some `ramp depth' $D_\text{ramp}$ (set to 5 in our evaluations), and half of the initial population grown randomly beginning from the root, with each node either being a leaf (terminal) node or interior (function) node with equal probability. Trees were depth limited to $D_\text{max}$, which due to limitations of the JVM, was set to $10$ for all runs. Parent selection was done with binary tournament selection.

Offspring were generated either via point mutation (with probability $p_m$) or crossover. Point mutation was done by replacing a (uniformly) randomly selected subtree of the parent with a new randomly grown tree, depth limited such that the offspring's size would respect $D_\text{max}$. Crossover was performed by randomly selecting a crossover point in each parent and generating new offspring by swapping the subtrees in each parent. In the event that one of the offspring resulting from crossover exceeded $D_\text{max}$, it was discarded.

In addition to the generated offspring, one elite clone was carried over from the previous generation to the next.

\subsection{Fitness}

The fitness of each algorithm was measured as the percentage of sheep that were successfully shepherded into the pen. The main disadvantage of this fitness measure was the difficulty in obtaining a fitness greater than zero for an initial random population. Another fitness measurement considered was the average distance of each sheep from the pen. However, this fitness measurement was often misleading in that sheep on the opposite side of the pen fence had very high fitness, despite require additional complex influences to force them around the fence into the pen. The average distance fitness was ultimately discarded in favor of the original fitness function.

An additional problem for both fitness functions was the high variance in the results of a simulation run due to the random initial conditions. For some simple test cases, the genetic algorithm could not make significant progress, due to the misleading fluctuations in the fitness.  These issues were partially mitigated by redefining the fitness to be the average fitness over $10$ simulations.


\section{Experimentation}\label{Experimentation}

\subsection{One Dog}

Initial experimentation took place under the conditions of a single dog attempting to shepherd $20$ sheep into the pen. The `dog AI' was given access to a limited set of only 4 parameters: its position ($x$ and $y$ coordinates) as well as those of the nearest free sheep.  This simplified scenario was targeted as an early goal because it restricts the complexity of the problem and provides good conditions for evolved behaviors. The experiment took place using a population size of $250$ over $220$ generations with the mutation rate $p_m$ was held at $0.05$.

\subsection{Multiple Dogs}

After achieving reasonable success in the single dog scenario, we extended the simulation to multiple dogs to determine if the dogs could develop team-like behaviors or if the genetic algorithm would just rediscover its successful single dog algorithm.  Simulations were run for 3 dogs attempting to herd the same $20$ sheep.  To encourage teamwork, the `dog AI' was given an expanded set of parameters: dog position, other dog positions, nearest free sheep, average sheep position, and a steering point.  The steering point was defined as the point a constant distance (10) behind the nearest free sheep position, in a line with it and the pen.  Note that because the dogs spawn in different locations, many of these parameters will be unique for each dog, allowing for divergent behavior.

The other simulation parameters were mostly carried over from the single dog scenario. In this case, a population size of $250$ was used and the algorithms were evolved for $100$ generations.


\section{Results}

As a baseline behavior from which evolved genetic algorithms could be compared, a handcrafted algorithm (referred to as `simple-dog') was created and evaluated. To maintain fairness, the handcrafted algorithm was limited to the same parameters and operators that were available  to the genetic algorithms. This algorithm was applied only to the single dog scenario over $10,000$ simulation runs, achieving an average fitness of $0.640 \pm 0.002$. For comparison, we also evaluated the effectiveness of `rand-dog', a dog that applies a (uniformly) random force vector to itself each time step, which achieved an average fitness of only $0.054 \pm 0.001$.

The earliest evolved algorithm that succeeded in shepherding any sheep into the pen we dubbed `split-dog', due to its tendency to shepherd the sheep by moving towards the flock at maximum velocity, splitting the flock apart. Remarkably, this algorithm evolved to split the flock apart in such a way that the sheep clustering force would cause the flock to reassemble itself \emph{in the pen} despite no further interactions with the dog (which tended to get `stuck' in the bottom right corner of the simulation environment). In one particularly successful simulation, `split-dog' achieved $15/20$ fitness; however, this relied on very specific environmental/initial conditions, and `split-dog' achieved very poor fitness on average.

\begin{figure}
\includegraphics[width=\linewidth]{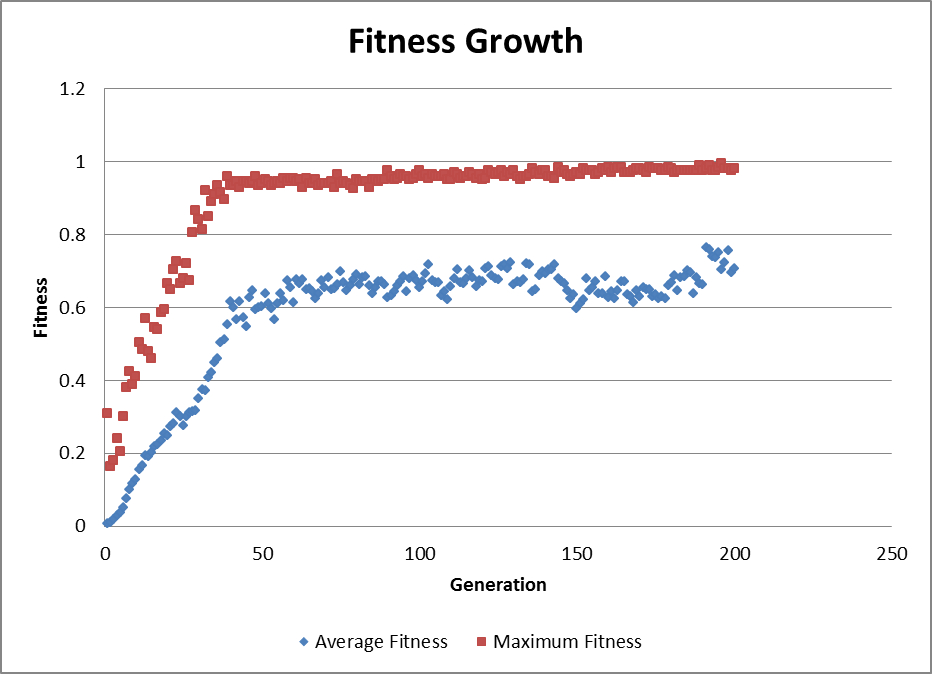}
\caption{Growth of Fitness Over Generation for a Single Dog, 150 Population Run}
\label{fig:fitnessgrowth}
\end{figure}

However, given sufficient generations, the genetic program produced an evolved behavior that performed well. Figure ~\ref{fig:fitnessgrowth} shows the growth of the maximum and average fitness values for the genetic algorithm population at each generation. The maximum fitness value converges close to perfect shepherding within 50 generations. Although the fitness growth can change between trials due to the nondeterministic nature of genetic programming and the simulation environment, typical results matched the growth curve seen here.

After $220$ generations, the best evolved algorithm had an average fitness of $0.743 \pm 0.003$, surpassing the `simple-dog' human-written algorithm.  Although the genetically evolved code is practically unreadable, the basic strategy can be inferred from the simulation visualization.  The dog waits for the sheep to cluster, then moves to the right of the field, pushing the sheep cluster to the left and underneath the pen.  The dog then moves back to the left and up, pushing the majority of the sheep into the pen.

The best evolved multiple dog algorithm had a fitness of $0.780 \pm 0.003$, slightly better than the single dog.  The dogs behaved in a similar manner, overlapping each other for most of the simulation, but near the end, as the sheep neared the pen, they would separate, some pushing from below, and others nearer the pen, pushing in from the sides to keep the sheep from escaping.  Although limited, this teamwork was enough to improve upon the results of the single dog algorithm.

\begin{figure}
\includegraphics[width=\linewidth]{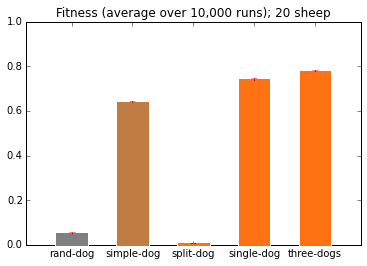}
\caption{Evaluated Fitness Over 10,000 Trials}
\label{fig:averagefitnessevolved}
\end{figure}

Due to the stochastic nature of the simulation (caused by the random initial placement of the sheep), we tested the evolved algorithms over a very large number of trials to confirm their effectiveness. The results in \ref{fig:averagefitnessevolved} show that the evolved single and three dog algorithms obtained a high average fitness over 10,000 trials. The `split-dog' serves as a comparison to the best evolved algorithms; `split-dog', evolved for only 10 generations actually fails to outperform `rand-dog'.

\subsection{Adaptability to Other Environmental Conditions}

One danger of genetic programming is the potential to overfit an algorithm. In some cases, an evolved algorithm could excel at completing its intended task but falter when confronted with slightly different task. To this end, the genetic algorithms were measured for fitness against tasks that were altered from the training task.

\begin{figure}[H]
\includegraphics[width=\linewidth]{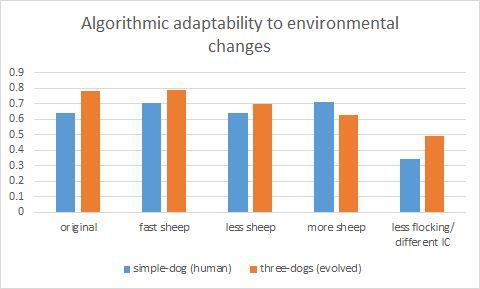}
\caption{Effect of Changing Simulation Parameters on Algorithmic Fitness}
\label{fig:adaptability}
\end{figure}

A small sample is shown in Figure \ref{fig:adaptability}, including runs with less sheep ($5$), more sheep ($100$), faster sheep ($v_s = 3$), and weaker sheep clustering force ($d_s = 5$) and different initial conditions (sheep spawn in the lower half of the field). Comparing the human written simple-dog algorithm to our evolved three-dog one, we see that both remain reasonably effective despite the changes to the environment. For these simple tests, the evolved algorithm appears as adaptive and general as the human written one.

\section{Discussion}

The experimental results demonstrates the potential of evolving complex behaviors through genetic programming. Although the operators and parameters made available to the genetic algorithms were quite limited, successful capabilities were evolved within 50 generations.

`split-dog' serves as an interesting demonstration of both genetic programming's ability to find novel solutions, and its tendency to produce drastically overfit solutions. However, given sufficient generations, the random initial sheep positions appeared to successfully force the genetic program to develop more \emph{generally} good solutions. The evolved solutions still perform very well, even under environmental conditions that the genetic program was not trained on. Unfortunately, however, the evolved genetic programs are generally very difficult, if not impossible for humans to understand.

There is ample room for further investigation using the presented framework. An interesting adaptation would be to increase the complexity of the problem by moving the pen away from the corner or limiting the vision of the dogs. In addition, the sheep could also be allowed evolve and adapt to the dog population.


\section{Implementation}

The entire genetic programming system and simulation environment for this project was implemented in Clojure (a dialect of Lisp) using only the standard library functions available with the language. The simulation visualizations were implemented in Java.

The full code repository (with links to videos of the simulations) is available online: \\
https://github.com/jtcb/flock


\end{document}